 \documentclass[final,5p,times,twocolumn,authoryear]{elsarticle}

\usepackage{amssymb}
\usepackage{lipsum}
\usepackage[utf8]{inputenc}
\usepackage{multirow}
\usepackage{algorithmic}
\usepackage{orcidlink}
\usepackage{subfigure}
\usepackage{amsmath} 
\usepackage{amsfonts,amssymb}  
\usepackage{colortbl}
\usepackage{makecell}
\usepackage{xcolor}
\usepackage{color}
\usepackage{threeparttable} 
\usepackage{xcolor}
\usepackage{graphicx}
\usepackage{mathrsfs}
\usepackage{stfloats}
\graphicspath{ {./figures/} }
\definecolor{mycolor}{RGB}{188,231,204}
\definecolor{mycolor2}{RGB}{228,238,188}
\definecolor{mycolor3}{RGB}{254, 248, 198}

\journal{Elsevier}

\begin{document}

\begin{frontmatter}



\title{Knowledge-Aware Mamba for Joint Change Detection and Classification from MODIS Times Series}


\author{Zhengsen Xu\textsuperscript{*}, Yimin Zhu\textsuperscript{*}, Zack Dewis, Mabel Heffring, Motasem Alkayid, Saeid Taleghanidoozdoozan, Lincoln Linlin Xu}
\affiliation{organization={Department of Geomatics Engineering, University of Calgary, Calgary, Canada}}

\begin{abstract}
Although change detection using MODIS time series is critical for environmental monitoring, it is a highly challenging task due to key MODIS difficulties, e.g., mixed pixels, spatial-spectral-temporal information coupling effect, and background class heterogeneity. This paper presents a novel knowledge-aware Mamba (KAMamba) for enhanced MODIS change detection, with the following contributions. First, to leverage knowledge regarding class transitions, we design a novel knowledge-driven transition-matrix-guided approach, leading to a knowledge-aware transition loss (KAT-loss) that can enhance detection accuracies. Second, to improve model constraints, a multi-task learning approach is designed, where three losses, i.e., pre-change classification loss (PreC-loss), post-change classification loss (PostC-loss), and change detection loss (Chg-loss) are used for improve model learning. Third, to disentangle information coupling in MODIS time series, novel spatial-spectral-temporal Mamba (SSTMamba) modules are designed. Last, to improve Mamba model efficiency and remove computational cost, a sparse and deformable Mamba (SDMamba) backbone is used in SSTMamba. On the MODIS time-series dataset for Saskatchewan, Canada, we evaluate the method on land-cover change detection and LULC classification; results show \~1.5–6\% gains in average F1 for change detection over baselines, and \~2\% improvements in OA, AA, and Kappa for LULC classification. 
\end{abstract}




\end{frontmatter}




\section{Introduction}



Human activities and natural processes (e.g., wildfires, global warming) are continually reshaping the Earth's land cover. The resulting albedo dynamics propagate through forest succession, the carbon cycle, and biodiversity, with wide-ranging ecological and societal impacts \citep{ZHAN2002336, PLANQUE201713, HUANG201478}. Detecting land-cover change (LCC) from time-series remote sensing data is therefore essential for climate monitoring, disaster response, and resource management. Among available sensors, MODIS provides dense temporal coverage at continental to global scales and has become a workhorse for dynamic, large-scale land-use/land-cover (LULC) change analysis \citep{FRIEDL2010168}.

\begin{figure}
    \centering
    \includegraphics[width=0.51\textwidth]{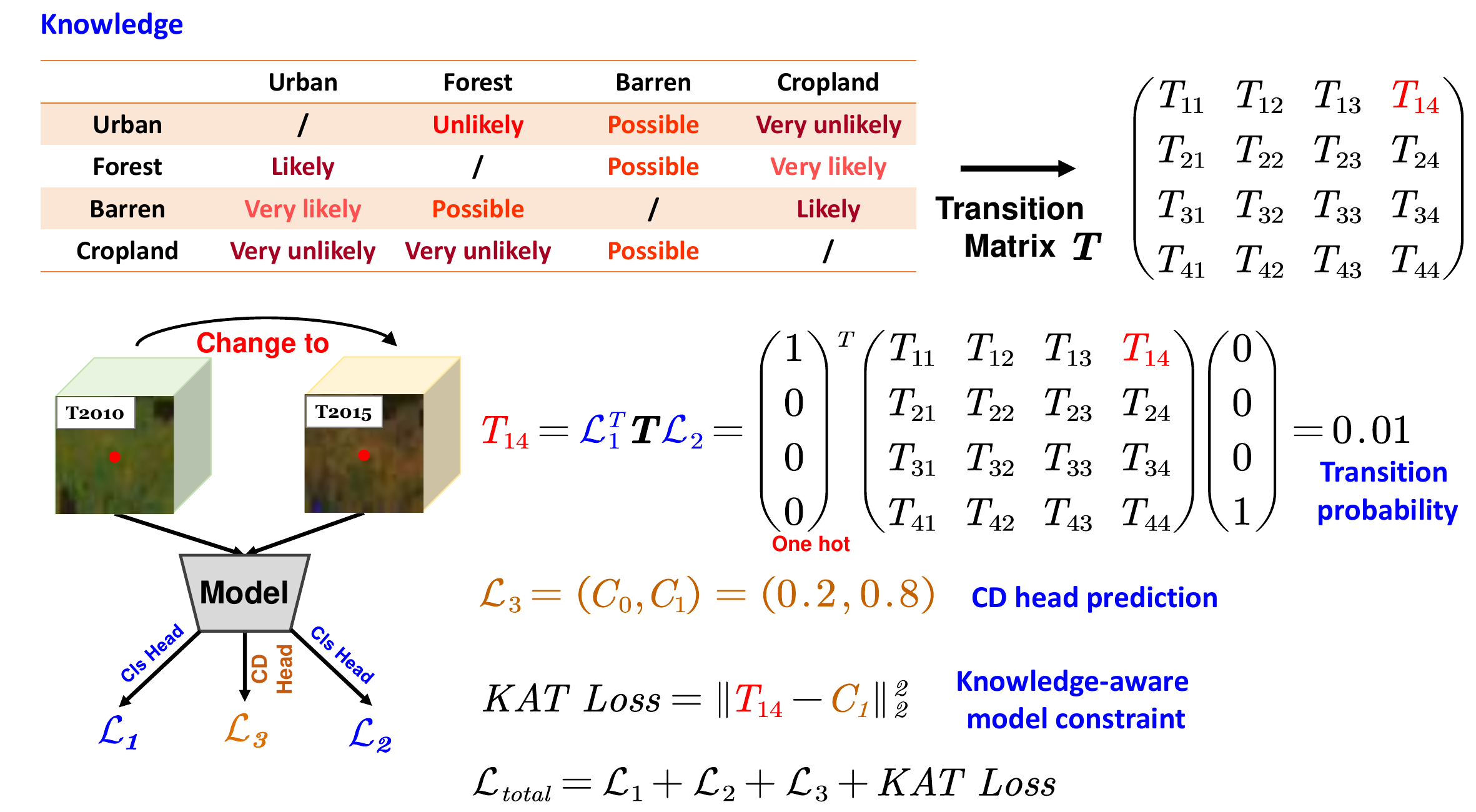}
    \caption{Illustration of the \textbf{K}nowledge-\textbf{A}ware \textbf{Mamba} approach (\textbf{KAMamba}). The transition matrix \(\textit{\textbf{T}}\) is built based on the existing prior knowledge. The model outputs three vectors, e.g., \(\mathcal{L}_1\), \(\mathcal{L}_2\), \(\mathcal{L}_3\), the classification probability of 2010 MODIS data, 2015 MODIS data, and the change probability. Take urban and cropland as an example, it is very unlikely that urban will be changed into cropland, with only 0.01 probability. If the change detection head outputs a possibility of 0.8 or higher, that means that the results of change detection are not consistent with the prior knowledge, leading to a high penalty for overestimated results, which is realized by the knowledge-aware transition loss (\textbf{KAT-Loss}). While in the existing methods, they do not include any connections or relationships to establish the correlation between different loss functions, they only use weight coefficients to balance the contribution of each loss, leading to unreasonable change directions.
     }
    \label{knowledge_KIT}
\end{figure}

A central difficulty in MODIS-based LCC is the \emph{heterogeneity of LULC transitions}: different source classes convert to different targets with markedly unequal probabilities, driven by human activities and environmental factors (\autoref{knowledge_KIT}) \citep{gaur2020spatio}. Yet many modern change-detection pipelines---including deep learning methods---implicitly treat transitions as class-agnostic or otherwise underutilize class-dependent priors \citep{8673805,9740122,bai2023deep}. This mismatch induces \emph{transition bias}, i.e., systematic misestimation of class-specific change likelihoods (e.g., implausible glacier$\!\to$cropland conversions), which undermines the reliability of outputs for environmental decision support. Although the remote-sensing literature offers rich spatiotemporal priors, these are frequently confined to preprocessing (sampling heuristics, band selection, vegetation indices) rather than being embedded within model architectures or training objectives.

Long, multi-band MODIS sequences introduce a second challenge: strong coupling among spatial, temporal, and spectral dimensions. While multi-year series capture seasonality and can mitigate cloud-related gaps, the resulting space--time--spectrum entanglement can mask class-specific cues and blur the boundary between genuine land-cover change and phenological variability \citep{chen2018mapping, song2021evaluation, sieber2022albedo, da2025occurrence}. This effect increases uncertainty and misclassification, particularly in heterogeneous or transitional landscapes \citep{li2022trend, 10416178}. Effective models must therefore disentangle and selectively attend to the most informative spatial, temporal, and spectral signals.

A third challenge arises from accuracy--efficiency trade-offs at MODIS scale. Convolutional networks are computationally efficient but limited by local receptive fields \citep{sun2025fourier}; Transformers provide global context but incur quadratic cost in sequence length due to self-attention \citep{10902569, 10944152}. Recent state-space models (e.g., Mamba) address sequence modeling with linear-time, input-adaptive updates to a fixed-dimensional recurrent state \citep{gu2023mamba, vision_mamba}. However, when state capacity or parameterization is insufficient, compressed summaries can attenuate the influence of early tokens; bidirectional or multi-pass scanning can restore long-range fidelity but increases constant factors in practice, partially offsetting efficiency gains. As a result, the net advantage of such sequence-efficient designs over attention-centric ones can be modest in end-to-end remote-sensing pipelines unless these trade-offs are carefully managed.

Guided by the above observations, we target three objectives for MODIS-based change detection: (i) encode class-conditional transition knowledge to reduce transition bias; (ii) disentangle spatial, temporal, and spectral factors to separate true change from seasonality; and (iii) sustain long-range modeling under realistic computational budgets. To this end, we propose KAMamba, a knowledge-aware, sequence-efficient Mamba framework tailored to long MODIS time series. KAMamba aligns training with ecological and biophysical priors so that predictions are regularized toward plausible transitions, and contributes four advances:

\begin{itemize}
  \item \textbf{Knowledge-Guided Transition Modeling.} We construct a knowledge-driven, asymmetric class transition matrix and embed it as a knowledge-aware transition loss (\textbf{KAT-loss}) to penalize implausible transitions and attenuate transition bias.
  \item \textbf{Multi-Task Constraint Learning.} To enhance class discriminability while preserving change semantics, we jointly optimize \textbf{PreC-loss}, \textbf{PostC-loss}, \textbf{Chg-loss}, and \textbf{KAT-loss}, enforcing consistency between pre-/post-change labels and the directionality of LULC transitions.
  \item \textbf{Spatial--Spectral--Temporal Disentanglement.} The \textbf{SST\-Mamba} module decouples MODIS signals via (i) group convolutions for temporal separation, (ii) spectral--spatial tokenization to isolate mixed-pixel signatures, and (iii) anchor- and self-attention–based token selection to improve token purity and class specificity.
  \item \textbf{Efficiency Optimization.} An \textbf{SDMamba} backbone reduces end-to-end compute relative to conventional scanning by combining dynamic token pruning with deformable state transitions, retaining long-range fidelity at lower cost.
\end{itemize}

The remainder of the paper is organized as follows. Section \ref{Related Works} talks about related work. Section \ref{methodology} illustrates the details of the proposed KAMamba approach. Section \ref{results} presents the experimental design and results. Section \ref{conclusion} concludes this study.

\section{Related work} \label{Related Works}




In this section, we first review recent advances in deep learning techniques for change detection using remote sensing imagery, with a particular focus on the evolution of model architectures and current research limitations. In addition, we discuss the types of prior knowledge utilized in change detection and the methods by which such knowledge is integrated into existing models.

\subsection{Deep Learning-based Change Detection}

Deep learning-based bi-temporal remote sensing image change detection has undergone three distinct evolutionary stages in terms of network architectures, similar to other computer vision tasks such as object detection and semantic segmentation: CNN-based, Transformer-based, and Mamba-based approaches. Despite the diversity in underlying architectures, most deep learning change detection models fundamentally adopt Siamese or pseudo-Siamese architectures for bi-temporal feature extraction and fusion \citep{peng2025deep}.

CNN-based change detection models represent the classical architectural paradigm, utilizing encoder-decoder structures for low-level and high-level local feature extraction and fusion. For instance, the SNUNet-CD model employs parameter-sharing convolutional decoders for bi-temporal feature extraction, incorporating dense skip connections and a Channel Attention Module for deep feature fusion \citep{9355573}. Bi-SRNet addresses the lack of integration between temporal and change branches by proposing two semantic reasoning blocks that enhance single-temporal semantic representation and bi-temporal semantic correlation reasoning \citep{9721305}. However, these change detection networks are fundamentally constrained by CNN's local feature extraction capabilities, lacking the ability to perceive long-range dependencies and model global semantic information effectively.

Compared to CNNs with local receptive fields, Transformers demonstrate superior capability in modeling global contextual relationships \citep{vaswani2017attention}. The fundamental Transformer architecture comprises multi-head self-attention mechanisms and multi-layer perceptrons. Although the basic model structures differ, Transformer-based change detection models maintain overall architectural similarities to CNN-based approaches, predominantly adopting Siamese or pseudo-Siamese frameworks. For example, TransUNetCD utilizes two CNN stems for bi-temporal feature extraction, concatenating the extracted low-level features along the channel dimension, followed by a Transformer bottleneck for long-range dependency modeling, and finally employing a CNN decoder with skip connections to recover detailed image features \citep{9761892}. However, the original Vision Transformer suffers from quadratic computational complexity growth with input size due to global self-attention. To address this limitation, the Swin Transformer, based on window partitioning and shifted window mechanisms, was proposed to reduce computational burden \citep{Liu_2021_ICCV}. Building upon this foundation, SwinSUNet designs a parallel dual-temporal encoder based on Swin Transformer, fusing outputs from each encoder layer through skip connections into the decoder, achieving more efficient long-range dependency modeling \citep{9736956}.

Although Transformers capture global dependencies more effectively than CNNs, the patchification process often leads to the neglect of local features. Consequently, hybrid models combining Transformers with CNNs have become prevalent in the change detection domain \citep{peng2025deep}. For instance, Feng et al. proposed ICIF-Net, which employs CNNs and Transformers to extract low-level features separately, subsequently integrating multi-scale features through intra-scale cross-interaction and inter-scale feature fusion to obtain fine-grained and object-level feature representations \citep{9759285}. However, Zhang et al. argued that simple combinations fail to consider the interactions between features extracted by different architectures. They proposed an Asymmetric Cross-Attention Hierarchical Network (ACAHNet) by combining CNNs and Transformers in series-parallel configurations, effectively reducing computational complexity while improving model performance \citep{10045704}. Similarly, Li et al. contended that existing simple concatenations of CNN and Transformer modules cannot capture multi-scale contextual information, resulting in limited change detection performance in small change regions. They proposed parallel branch ConvTrans modules for local and global feature extraction to better distinguish between changed and unchanged regions \citep{10114976}.

The aforementioned developmental trends reveal that deep learning-based change detection networks have become increasingly sophisticated, with the introduction of Transformers significantly elevating computational demands. Consequently, Mamba \citep{gu2024mamba}, based on state space models, has attracted considerable attention in the change detection field due to its linear computational complexity, parameter-efficient design, and effective long-range dependency modeling capabilities. For example, Chen et al. constructed three change detection models using weight-sharing multi-layer Mamba encoder-decoder blocks, exploring Mamba's effectiveness compared to CNNs and Transformers in binary change detection, semantic change detection, and building damage assessment \citep{10565926}. Experimental results demonstrated that simply designed Mamba networks could surpass current sophisticated CNN and Transformer-based models across these three tasks. Additionally, Paranjape et al. designed encoders comprising hierarchical Mamba modules, employing two weight-sharing encoders for bi-temporal feature extraction and modeling feature differences through a Difference Module, finally obtaining change detection results through a decoder consisting of multi-layer channel-average Mamba modules \citep{10944152}. Compared to self-supervised or diffusion methods, Mamba models require shorter training times while demonstrating significant performance advantages. To further enhance Mamba's fine-detail perception capabilities, Chen et al. combined CNNs with Mamba to design Scaled Residual ConvMamba (SRCM) modules, addressing the local detail information loss problem in current Mamba-based approaches for dense prediction tasks \citep{10902569}. Furthermore, this research proposed an Adaptive Global-Local Guided Fusion (AGLGF) module for more effective feature interaction, thereby obtaining more discriminative change features.

\subsection{Knowledge Informed Change Detection}

Current remote sensing image change detection deep learning models typically build upon established concepts from computer vision, particularly semantic segmentation. This approach has unfortunately led to a neglect of domain-specific prior knowledge in remote sensing \citep{peng2025deep}. Consequently, many developed models suffer from limited transferability and interpretability \citep{Deng2020, Moskolaï2021}. To address these limitations, researchers have begun proposing knowledge-informed change detection models.

For instance, \cite{9699382} developed a graph-based network that extracts shared structural knowledge from labeled datasets, which then guides the optimization of semi-supervised SAR image change detection models. Building on this knowledge transfer concept, \citep{10373904} introduced a spectral knowledge transfer (SKT) framework that initially trains an autoencoder on paired RGB and hyperspectral (HSI) natural images to learn bidirectional spectral mapping. The framework's decoder is subsequently calibrated using real multispectral remote sensing images, enabling the generation of pseudo-HSI representations from multispectral inputs. During actual change detection, the pretrained encoder remains frozen while the calibrated decoder enhances the spectral richness of multispectral imagery, resulting in improved detection performance.

However, despite their improved accuracy, these methods still face limitations in terms of knowledge interpretability. To address this issue, \cite{10109790} incorporated four remote sensing indices into a self-supervised learning framework, enabling the model to distinguish actual land-cover changes from noise by leveraging physically meaningful features. Similarly, \cite{10855633} proposed a statistical color alignment strategy that adjusts RGB channels of bitemporal images based on channel-wise mean and variance, effectively mitigating spectral inconsistencies caused by seasonal variations or sensor differences and thereby improving the reliability of change detection.

Integrating domain knowledge into deep learning networks not only enhances predictive accuracy but also reduces sensitivity to limited or noisy data \citep{Karniadakis2021}. Nevertheless, most existing change detection models still lack systematic mechanisms for embedding remote sensing–specific knowledge \citep{Karniadakis2021}. Although recent studies have introduced knowledge-guided frameworks \citep{9699382, 10109790, 10373904}, there is still no standardized methodology for fusing multi-temporal domain knowledge in a unified manner. This gap persists even though evidence consistently shows that knowledge-informed models require less training data \citep{9429985}, yield superior predictive performance, and maintain stronger physical consistency \citep{Moskolaï2021, Deng2020}. As emphasized by \citep{peng2025deep}, the field urgently needs principled approaches to embed spectral indices, temporal patterns, and geophysical principles directly into end-to-end deep learning pipelines, moving beyond the current over-reliance on computer vision paradigms.

\section{Methodology} \label{methodology}

\subsection{Preliminaries}
State-Space Model (SSM) and Mamba \citep{gu2024mamba} are inspired by linear
time-invariant systems, which can effectively model long-range dependencies by conceptualizing continuous systems that map 1-D input sequences \(x(t) \in \mathcal{R}^{M_1}\) to a response \(y(t) \in \mathcal{R}^{M_2}\) through a hidden state
\(h(t) \in \mathcal{R}^N\).
\begin{align}
    \begin{aligned}
        &h'(t) = \mathbf{A}h(t) + \mathbf{B}x(t) \\
        &y(t) = \mathbf{C}h'(t)
    \end{aligned}
\end{align}
where \(\mathbf{A} \in \mathbb{R}^{N \times N}\) represents the state matrix and \(\mathbf{B} \in \mathbb{R}^{N \times M_1}\)
, and \(\mathbf{C} \in \mathbb{R}^{M_2 \times N}\) are the projection parameters.

When continuous systems are further discretized, a common discretization method is the zero-order hold, which leads to
\begin{align}
    \begin{aligned}
        &\overline{\mathbf{A}} = \exp(\Delta \mathbf{A}) \\
        &\overline{\mathbf{B}} = (\Delta \mathbf{A})^{-1} (\exp (\Delta \mathbf{A}) - \mathbf{I}) \Delta \mathbf{B}
    \end{aligned}
    \label{EQ2}
\end{align}
where, \(\Delta\) is the timescale parameter, representing the sampling step. After that, the discretization of \autoref{EQ2} can be formulated as:
\begin{align}
    \begin{aligned}
        &h'_t = \overline{\mathbf{A}}h_{t-1} + \overline{\mathbf{B}} x_t \\
        &y_t = \mathbf{C}h_t
    \end{aligned}
\end{align}

It can also be expressed in a convolutional form to accelerate
computation:
\begin{align}
    \begin{aligned}
        &\overline{\mathbf{K}} = (\mathbf{C} \overline{\mathbf{B}}, \mathbf{C} \overline{\mathbf{A}} \overline{\mathbf{B}}, ..., \mathbf{C} \overline{\mathbf{A}}^{L-1} \overline{\mathbf{B}}) \\
        &y = x * \overline{\mathbf{K}}
    \end{aligned}
\end{align}
where \(\overline{\mathbf{K}} \in \mathbb{R}^L\) is a structured convolutional kernel and \(*\) denotes a convolutional operation.

\subsection{Problem Statement}

Recent selective state-space models, such as Mamba \citep{gu2024mamba}, maintain a fixed-dimensional recurrent state and update it in an input-adaptive manner, thereby enabling linear-time modeling of long sequences. However, as sequence length increases, information from early tokens can be attenuated when state capacity or parameterization is constrained. Consequently, input design—particularly the definition, tokenization, and compression of sequences—becomes critical. Motivated by this, we adopt a cascaded Mamba-based architecture with sparse and deformable tokens to capture long-range spatial–spectral–temporal dependencies in MODIS time series.

We consider a multi-task setting comprising annual land-cover classification and change detection. Prior work \citep{10565926,10959091,SHEN2025104409} often optimizes these outputs independently and lacks explicit constraints that align per-year class posteriors with the change map. In contrast, we introduce a class-conditioned transition prior represented by a matrix \(\mathbf{P}\in[0,1]^{K\times K}\), which encodes plausible LULC transitions. This prior couples the classification vectors and the change logits via a transition-regularized objective, aligning predictions with established class-dependent asymmetries in land-cover dynamics, mitigating transition bias, and improving overall consistency.


\subsection{Overview}

In terms of data format, samples have the structure
\(
\mathcal{D}
=\big\{\big(\mathbf{X}^{2010}_j,\mathbf{X}^{2015}_j,\mathbf{y}^{2010}_j,\mathbf{y}^{2015}_j,\mathbf{C}^{\mathrm{cd}}_j\big)\big\}_{j=1}^{N}.
\)
Each sample comprises cropped MODIS patches from 2010 and 2015,
\(\mathbf{X}^{2010}_j,\mathbf{X}^{2015}_j \in \mathbb{R}^{C\times T\times H\times W}\),
with \(C=6\) bands, including NDVI, EVI, MODIS band 1, 2, 3, and 7, \(T=23\) temporal frames, and spatial size \(H=W=13\).
Thus, each patch contains \(C\times T=138\) channels per pixel.
The corresponding labels include per-year land-cover annotations
\(\mathbf{y}^{2010}_j\) and \(\mathbf{y}^{2015}_j\), as well as a change-detection label
\(\mathbf{C}^{\mathrm{cd}}_j\).

\begin{figure}[!t]
  \centering
  \includegraphics[width=\columnwidth]{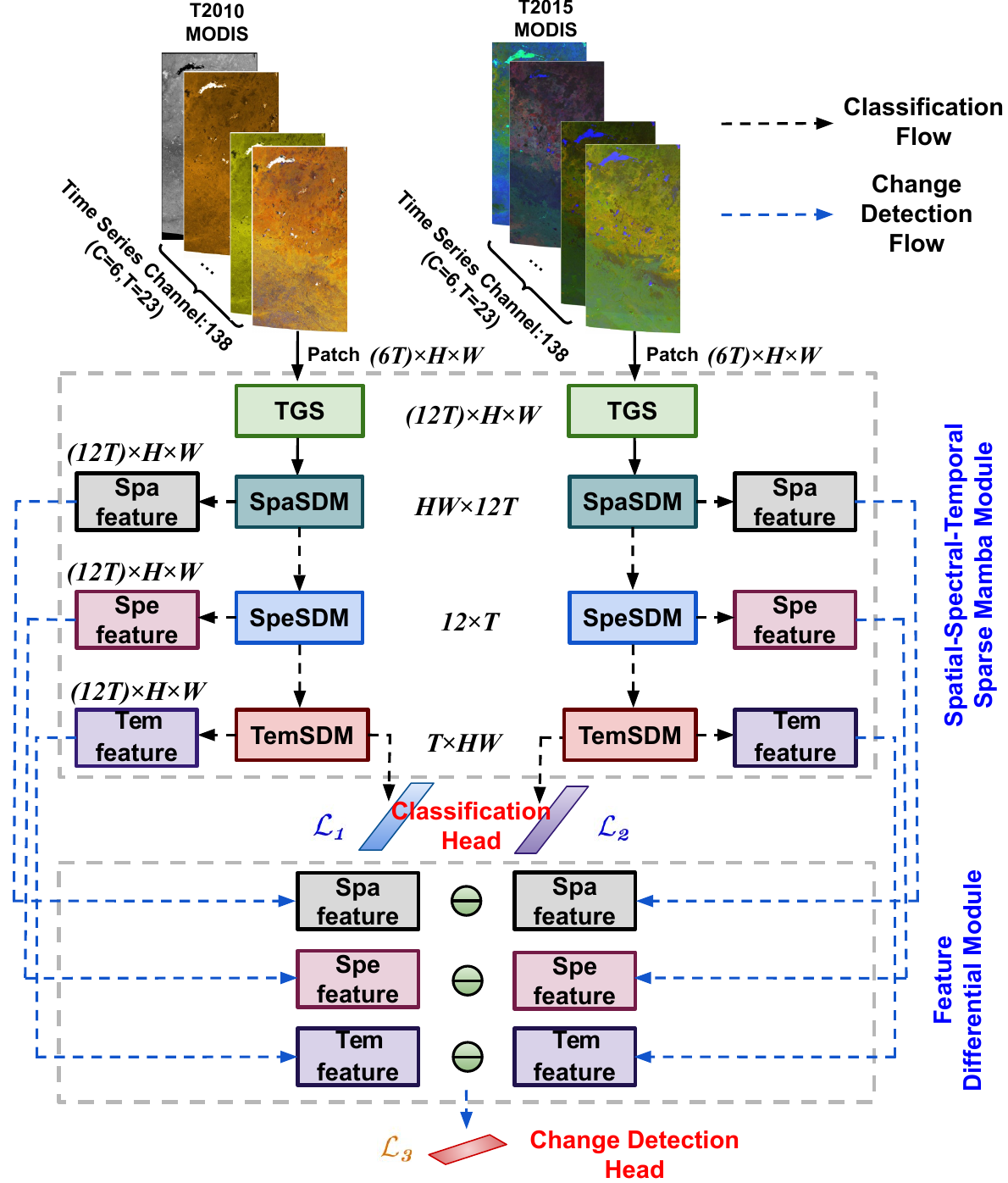} 
  \caption{Overview of our proposed Knowledge-aware Mamba for MODIS time series change detection. It is a sequential model that contains the Spatial-Spectral-Temporal Sparse Module (SST-SM) \textbf{without weight sharing} and the Feature Differential Module (FDiffM). In SST-SM, Temporal Group Stem (\textbf{TSG}), Spatial Sparse Deformable Mamba (S\textbf{paSDM}), Spectral Sparse Deformable Mamba (\textbf{SpeSDM}), and Temporal Sparse Deformable Mamba (\textbf{TemSDM}) are included; the corresponding output features are used for building the input feature of \textbf{FDiffM}. The differential features are concatenated and then fed to the change detection head.}
  \label{model_overview}
\end{figure}

Second, the proposed KAMamba is a sequential multi-task model without weight sharing, independently extracting the  MODIS spatial-spectral-temporal feature from two different time series, as shown in \autoref{model_overview}. The cropped MODIS image patch is fed into the classification flow, as shown in the gray dashed line, which can be formulated as follows:
\newcommand{\TemSDM}{\mathop{\mathrm{TemSDM}}\nolimits}
\newcommand{\SpeSDM}{\mathop{\mathrm{SpeSDM}}\nolimits}
\newcommand{\SpaSDM}{\mathop{\mathrm{SpaSDM}}\nolimits}

\begin{align}
\begin{aligned}
\mathbf{Z}_{j}^{2010} &= \TemSDM(\SpeSDM(\SpaSDM(\mathbf{X}_{j}^{2010}))\bigr) \\
\mathbf{Z}_{j}^{2015} &= \TemSDM(\SpeSDM(\SpaSDM(\mathbf{X}_{j}^{2015}))\bigr)
\end{aligned}
\end{align}

For the change detection flow, blue dashed lines in ~\autoref{model_overview}, the extracted features of each sparse demormable Mamba block \(\mathscr{A}, \mathscr{B}, \mathscr{C}\) at the same layer are used as input of the Feature Differential Module, as follows:
\begin{align}
    \begin{aligned}
        \mathcal{F}_{j}^{spa} = |\mathscr{A}_{j}^{2010} - \mathscr{A}_j^{2015}| \\ 
        \mathcal{F}_{j}^{spe} = |\mathscr{B}_{j}^{2010} - \mathscr{B}_j^{2015}| \\ 
        \mathcal{F}_{j}^{tem} = |\mathscr{C}_{j}^{2010} - \mathscr{C}_j^{2015}| \\ 
    \end{aligned}
\end{align}
\(\mathcal{F}_{j}^{spa}\), \(\mathcal{F}_{j}^{spe}\), and \(\mathcal{F}_{j}^{tem}\) are used to perform change detection head after doing the average pooling on the spatial dimension.

\subsection{Spatial-Spectral-Temporal Mamba (SSTMamba)}

\begin{figure*}[]
    \centering
    \includegraphics[width=\textwidth]{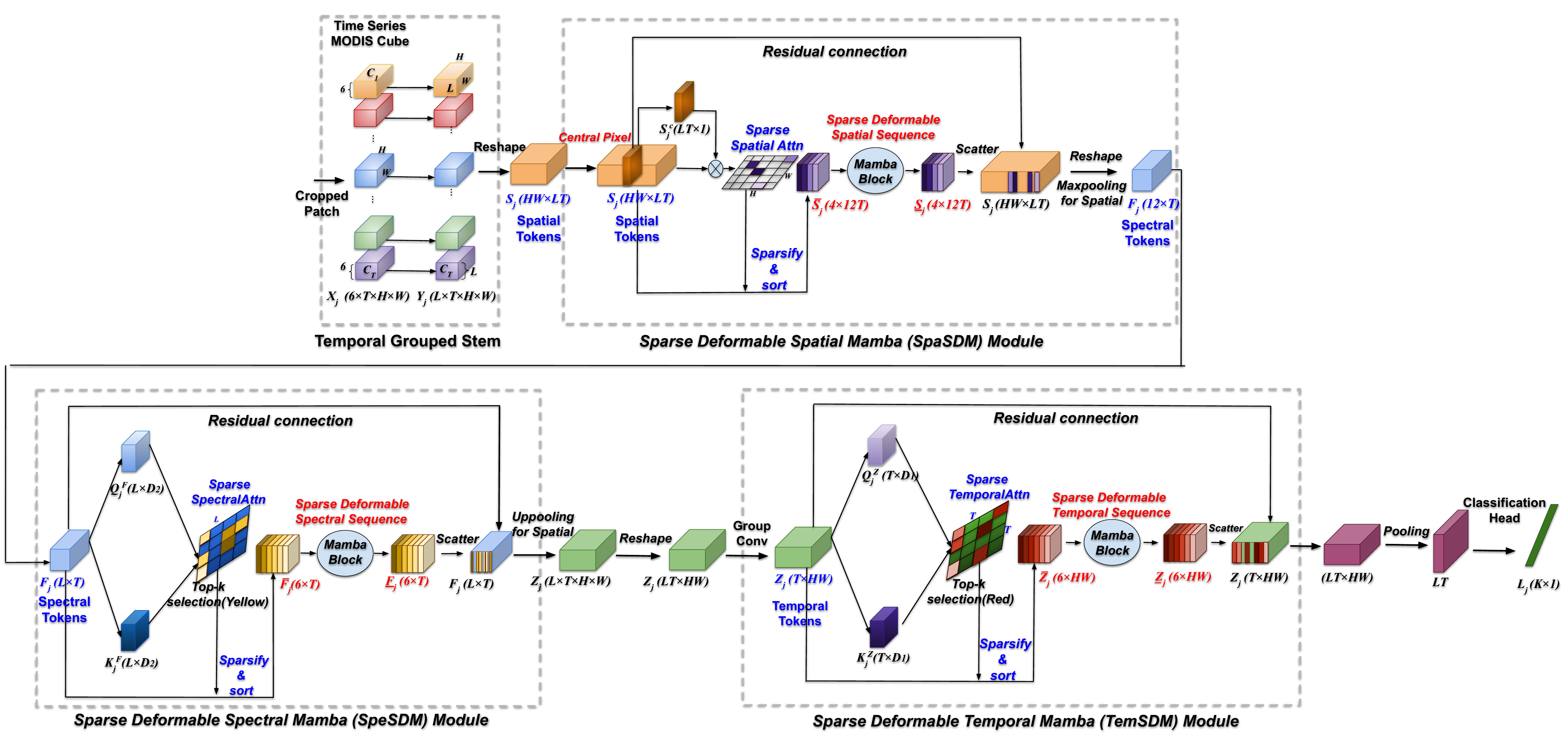}
    \caption{Overview of the Spatial-Spectral-Temporal Sparse Deformable Mamba.After TGS module, the feature is reshaped into \(HW \times LT\) as spatial tokens and then fed into SDSpaM. Here, we set \(L=12\). The output of SDSpaM is reshaped into \(L \times T\) after doing the maxpooling on the spatial dimension as the input of SDSpeM. For SDTemM, the input is being the shape of \(T \times HW \). All the spatial, spectral, and temporal modules use the attention mechanism to select dynamically the most informative tokens to reduce the redundancy.}
    \label{TGS}
\end{figure*}

\subsubsection{\textbf{Temporal Grouped STEM (TGS)}}
As illustrated in \autoref{TGS}, the TGS layer is first used to extract low-level spatial information while preserving temporal structure information using a depthwise convolution layer (DWConv). It separates the input data cube $\mathbf{X}_j \in \mathbb{R}^{B \times (6 \times T) \times H \times W}$ into $T$ groups along the second dimension in a mini-batch $B$. Each temporal group with the size of $(B \times 6 \times H \times W)$ is processed individually by a depth-wise convolutional layer with kernel size $3 \times 3$. 

The equation for the TGS layer is expressed as:
\begin{align}
\begin{aligned}
    \mathbf{Y}_j = \text{GELU}(\text{BN}(\text{DWConv}(\mathbf{X}_j)))
\end{aligned}
\end{align}
where $\text{GELU}$ represents Gaussian Error Linear Units and $\text{BN}$ represents Batch Normalization. The global batch normalization ensures consistent distribution across each temporal group.


\subsubsection{\textbf{Sparse Deformable Spatial Mamba (SpaSDM)}}
The SpaSDM module processes the reshaped output $\mathbf{Y}_j \in \mathbb{R}^{LT\times H\times W}$ from the Temporal Grouped Stem by first flattening it into spatial tokens $\mathbf{S}_j \in \mathbb{R}^{HW \times LT}$, where \(L=12\). As shown in \autoref{TGS}, the module performs four key processing stages. First, anchor-guided sparsification uses the central pixel of each patch $\mathbf{S}_j^c \in \mathbb{R}^{LT \times 1}$ as a reference anchor, computing cosine similarity for each spatial position $i$ through 
\begin{align}
    \begin{aligned}
        \mathcal{A}_i = \arccos\left(\frac{\mathbf{S}_i^T \mathbf{S}_j^c}{\|\mathbf{S}_i\| \|\mathbf{S}_j^c\|}\right)
    \end{aligned}
\end{align}
where $\mathcal{A} \in \mathbb{R}^{H \times W}$ defines the full spatial attention map. 

Next, token pruning sorts tokens by $\mathcal{A}_i$ values, retaining only the top-$k$ ($k = \lambda HW$, $\lambda=0.3$) most similar tokens to generate a condensed sequence $\overline{\mathbf{S}}_j \in \mathbb{R}^{k \times CT}$ that preserves spatially coherent regions while discarding heterogeneous pixels.
The deformable Mamba processing stage then handles the pruned tokens through a Mamba block to capture long-range dependencies, with the output being scattered back to original spatial positions and residually combined with the input spatial tokens. Finally, spatial max pooling operates across the $HW$ dimension, followed by reshaping to produce spectral tokens $\mathbf{F}_j \in \mathbb{R}^{12 \times T}$. This pooling stage serves dual purposes: preserving the most salient spectral features across spatial locations while simultaneously reducing dimensionality to enable efficient spectral processing in subsequent modules. 

\subsubsection{\textbf{Spectral Sparse Deformable Mamba (SDSpeM)}}
After obtaining the most salient spectral feature and forming the spectral tokens \(\mathbf{F}_j \in \mathbb{R}^{L \times T}\), it will be processed by the common attention mechanism after the normalization and projection layer, the the full spectral attention map \(\mathbf{SpeAM}\) can be expressed as follows:
\begin{align}
    \begin{aligned}
        \mathbf{SpeAM}\ = Attention(\mathcal{Q}_j^F, \mathcal{K}_j^F) = \sigma(\frac{\mathcal{Q}_j^F {\mathcal{K}_j^F}^T}{\sqrt{D}_2})
    \end{aligned}
    \label{EQ_sparseattn}
\end{align}
where \(\mathcal{Q}_j = F_j \mathcal{W}^{\mathcal{Q}_j^F} \in \mathbb{R}^{L\times D_2} \), \(\mathcal{K}_j^F = F_j \mathcal{W}^{\mathcal{K}_j^F} \in \mathbb{R}^{L\times D}\) are queries, keys of spectral tokens. \(\mathcal{W}^{\mathcal{Q}_j^F}\) and \(\mathcal{W}^{\mathcal{K}_j^F}\)
represent the projection weights for \(\mathcal{Q}_j^F\), \(\mathcal{K}_j^F\), \(D_2\) represents the hidden dimension. \(\sigma\) indicates Softmax function. \(\mathbf{SpeAM}\) \(\in \mathbb{R}^{L \times L}\).   

To identify informative spectral tokens, we sparsify the spectral attention
\(\mathbf{SpeAM}\in\mathbb{R}^{L\times L}\) by ranking rows and columns separately and
taking the union of the selected indices. Let \(k=\lfloor \lambda\cdot L\rfloor\) with \(\lambda=0.3\).
Define
\begin{align}
\begin{aligned}
    &s^{\mathrm{row}}_i \;=\; \tfrac{1}{L}\sum_{t=1}^{L}\mathbf{SpeAM}_{i,t}\\
    &s^{\mathrm{col}}_j \;=\; \tfrac{1}{L}\sum_{t=1}^{L}\mathbf{SpeAM}_{t,j}
    \label{eq:rowcol-mean}
\end{aligned}
\end{align}
and the Top-\(k\) index sets
\begin{align}
\begin{aligned}
    &S_{\mathrm{row}} \;=\; \mathrm{TopK}_k\big(\{s^{\mathrm{row}}_i\}_{i=1}^{L}\big) \\
    &S_{\mathrm{col}} \;=\; \mathrm{TopK}_k\big(\{s^{\mathrm{col}}_j\}_{j=1}^{L}\big) \label{eq:rowcol-topk}
\end{aligned}
\end{align}

The final spectral token set is the union
\begin{align}
S \;=\; S_{\mathrm{row}} \,\cup\, S_{\mathrm{col}}. \label{eq:union}
\end{align}
The spectral mask and sparse attention are given by
\begin{align}
\begin{aligned}
&M^{\mathrm{spe}}_{i,j} \;=\; \mathbb{I}\big(i\in S \,\lor\, j\in S\big)\\
&\textit{SparseSpectralAttn} \;=\; M^{\mathrm{spe}}\odot \mathbf{SpeAM}, \label{eq:sparse-attn}
\end{aligned}
\end{align}
where \(\mathbb{I}(\cdot)\) is the indicator function (1 if the condition holds, 0 otherwise), 
\(\mathrm{TopK}_k(\cdot)\) returns the indices of the \(k\) largest entries of its vector argument,
and \(\odot\) denotes the Hadamard product.

The salient and condensed sequence \(\overline{\mathbf{F}}_j \in \mathbb{R}^{L\lambda \times T}\) is then obtained and fed into the Spectral Mamba block to learn the sequential relationship.  The output is scattered back to the original spectral positions and residually combined with the input spectral tokens. An up-pooling operation is used to recover the spatial size and is residually combined with the output of SDSpaM, followed by reshaping to produce the temporal tokens \(\mathbf{Z}_j \in \mathbb{R}^{LT \times HW}\).
\subsubsection{\textbf{Sparse Deformable Temporal Mamba (TemSDM)}}
The Temporal Mamba block is designed to capture long-range temporal dependencies by treating the temporal steps as a sequence. Since \(\mathbf{Z}_j\) still mix spectral together with temporal, a group convolution layer with total \(T\) groups is used to compress the channel, and form temporal tokens \(\mathbf{Z}_j \in \mathbb{R}^{T \times HW}\), which only use one salient channel to represent the spectral feature. Similar to SpeDM, the temporal attention map \(\mathbf{TemAM}\) is also calculated as :
\begin{align}
    \begin{aligned}
        \mathbf{TemAM} = Attention(\mathcal{Q}_j^Z, \mathcal{K}_j^Z) = \sigma(\frac{\mathcal{Q}_j^Z {\mathcal{K}_j^Z}^T}{\sqrt{D}_1})
    \end{aligned}
\end{align}
where \(\mathcal{Q}_j = \mathbf{Z}_j \mathcal{W}^{\mathcal{Q}_j^Z} \in \mathbb{R}^{T\times D_1} \), \(\mathcal{K}_j^Z = \mathbf{Z}_j \mathcal{W}^{\mathcal{K}_j^Z} \in \mathbb{R}^{T\times D1}\) are queries, keys of temporal tokens. \(\mathcal{W}^{\mathcal{Q}_j^Z}\) and \(\mathcal{W}^{\mathcal{K}_j^Z}\)
represent the projection weights for \(\mathcal{Q}_j^Z\), \(\mathcal{K}_j^Z\), \(D_1\) represents the hidden dimension. \(\sigma\) indicates Softmax function. \(\mathbf{TemAM}\) \(\in \mathbb{R}^{T \times T}\). The sparse temporal attention is written as follows:
\begin{align}
    \begin{aligned}
        M_{i,:}^{temp} = \mathbb{I}(TopK(\mathbb{E(}\mathbf{r}_i), \lfloor \lambda \times L \rfloor)) \\
        M_{:,j}^{temp} = \mathbb{I}(TopK(\mathbb{E}(\mathbf{a}_j), \lfloor \lambda \times L \rfloor)) \\
    \end{aligned}
\end{align}

Using this masking matrix \(M^{temp}\), the sparse temporal attention matrix is formed for updating the values of attention mechanism as well as the condensed input of Mamba block \(\overline{\mathbf{Z}}_j\). After scattering to the original shape and residual connection, the final classification prediction is performed through two linear layers.

\subsection{Spatial-Spectral-Temporal Feature Differential Module}
\label{change_head}
To form the final change detection prediction, a feature differential module is proposed to fuse all he spatial, spectral, and temporal features. First, the absolute error at the same stage is first calculated, and then concatenated along the channel dimension. An averaging pooling is used to aggregate the spatial feature, as the following equation:
\begin{align}
    \begin{aligned}
        \mathcal{F}_j = \text{cat}[\text{AvgPool}(\mathcal{F}^{spa}_j), \text{AvgPool}(\mathcal{F}^{spe}_j), \text{AvgPool}(\mathcal{F}^{tem}_j)]
    \end{aligned}
\end{align}

The change detection head is defined by two linear layers and an activation function as the following equation:
\begin{align}
    \begin{aligned}
        C_j = \text{FFN}(\mathcal{F}_j)
    \end{aligned}
\end{align}
where \(C_j\) is the change probability with the shape of \(2 \times 1\).
\subsection{Multi-task learning for joint change detection and classification}

\subsubsection{Pre-Change Classification HEAD}
Given the output of \textit{TemSDM} \(Z_j^{2010} \in \mathbb{R}^{B \times LT \times H \times W} \) on 2010 MODIS image data, the averaged feature is used to perform the auxiliary classification task, which is defined by a feedforward with two linear layers and a GELU activation function:
\begin{align}
    \begin{aligned}
        L_j^{2010} = \text{FFN}(\text{AvgPool}(Z_j^{2010}))
    \end{aligned}
\end{align}
The cross-entropy loss is adopted for updating the parameter, which can be written as:
\begin{align}
    \begin{aligned}
        \mathcal{L}_{1} &= \frac{1}{N} \underset{i=1}{\overset{N}{\varSigma}} \underset{k=1}{\overset{K}{\varSigma}} [
        y_{i}(c) \log L_j^{2010}(c) \\
        &- (1-y_{i}(c)) \log (1-L_j^{2010}(c))]
    \end{aligned}
\end{align}
where \(y_i\) denotes the class label of the \textit{i}th cube in training set with total number of \(N\). \(L_{j}^{2010}\) is the probability that the model predicts that sample \(i\) belongs to the category \(k\).

\subsubsection{Post-Change Classification HEAD}
For post MODIS image in 2015, a feedforward with two linear layers and a GELU activation function is also used to get the predicted vector and a cross-entropy loss is adopted as well. 
\begin{align}
    \begin{aligned}
        &L_j^{2015} = \text{FFN}(\text{AvgPool}(Z_j^{2015})) \\ 
        &\mathcal{L}_{2} = \frac{1}{N} \underset{i=1}{\overset{N}{\varSigma}} \underset{k=1}{\overset{K}{\varSigma}} [
        y_{i}(c) \log L_j^{2015}(c) \\
        &- (1-y_{i}(c)) \log (1-L_j^{2015}(c))]
    \end{aligned}
\end{align}

\subsubsection{Change Detection HEAD}
As mentioned in Section \ref{change_head}, we use binary cross-entropy to measure the difference between the predicted label and the ground truth, which can be written as follows:
\begin{align}
    \begin{aligned}
        \mathcal{L}_3 = \frac{1}{N} \underset{i=1}{\overset{N}{\varSigma}} \underset{l=1}{\overset{L}{\varSigma}} Y_{i}(l)log(C_{i}(l))
    \end{aligned}
\end{align}
where \(Y_i\) is the true change detection label, which denotes the label of \textit{i}th sample on the category l (0 for unchanged, 1 for changed).

\begin{figure}
    \centering
    \includegraphics[width=0.49\textwidth]{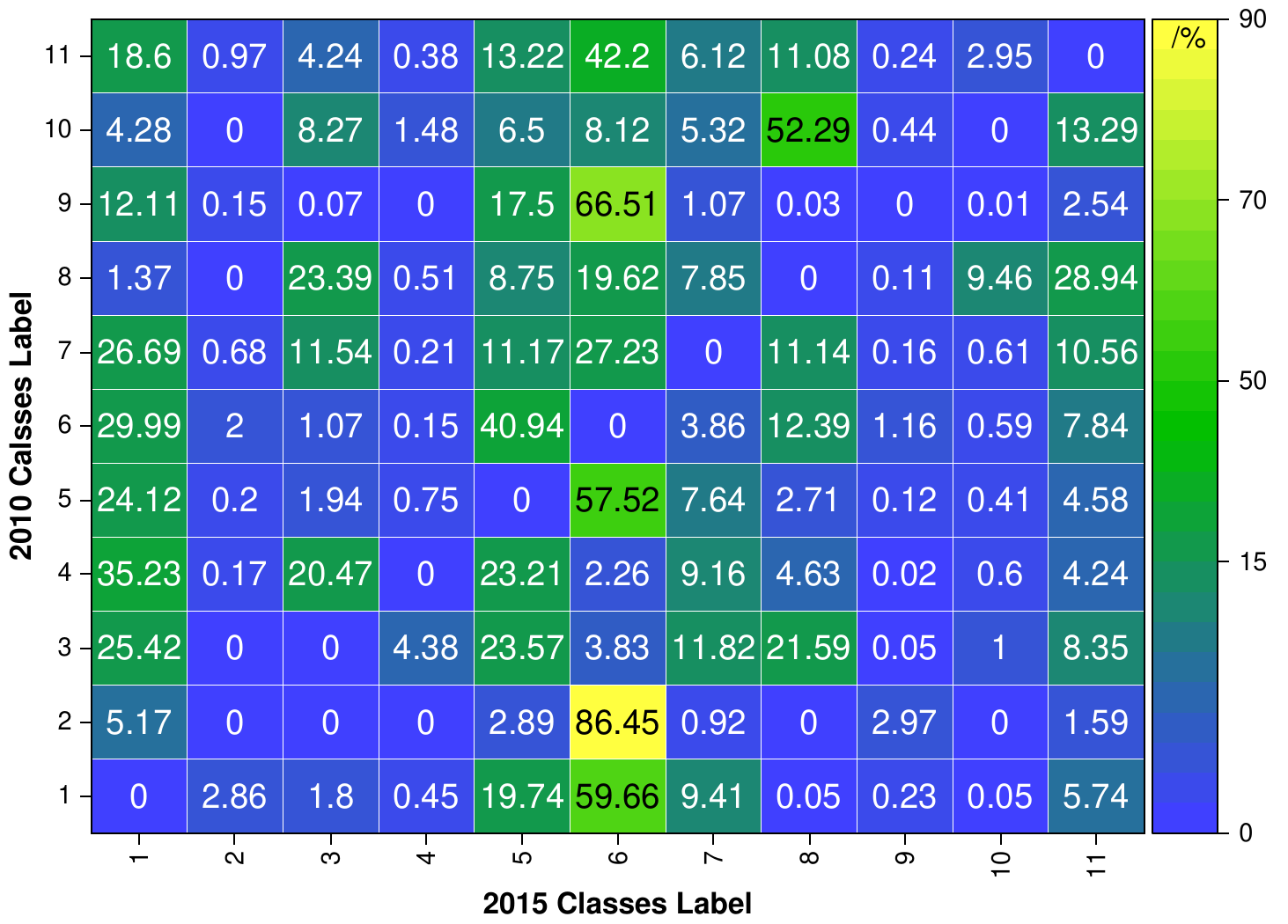}
    \caption{The transition matrix for building the knowledge-aware loss KAT-loss.}
    \label{Transition matrix}
\end{figure}

\subsection{Contrastive Learning Loss (CL Loss)}

Directly learning the binary classification boundaries of "changed/unchanged" is susceptible to the influence of imbalanced data, leading to model bias on the majority class. While, Contrastive Learning \citep{SimCLR} can bring similar samples (sharing the same label on MODIS 2010 and MODIS 2015) closer (unchanged pairs) and pushing different samples (changed pixels) away, a decoupled representation is constructed in the feature space to avoid the decision boundary being dominated by the majority class, letting model learns more general features from limited minority class samples. The contrastive learning loss is formulated as follows:
\begin{align}
    \begin{aligned}
        \mathcal{L}_{CL} = \log \frac{\exp(\mathcal{S}_{ij})}{\Sigma_{j \neq i} \exp(\mathcal{S}_{ij}) + \exp(\mathcal{S}_{ii} ) \times (1 - \mathbf{M}_{align}^j)}
    \end{aligned}
\end{align}
where \(\mathcal{S}\) is the similarity matrix calculated on the normalized classification prediction vector. \(\mathbf{M}_{align} = \mathbb{I}(y^{2010}_j, y^{2015}_j)\).
\subsection{Knowledge-aware Class Transition Loss (KAT-loss)}
Before building the KAT-loss, the prior knowledge, e.g., the transition matrix \(\textit{\textbf{T}}\), needs to be defined. Based on the change map, we can figure out the specific quantity of how one type of land cover has transformed into another. From the change to perspective, we can build a transition matrix \(\textit{\textbf{T}}\), which records the total number of transition categories for all the 11 land cover types. As shown in \autoref{Transition matrix}, land-cover transitions are markedly non-uniform: each class tends to convert to one (occasionally two) specific target classes with substantially higher probability than to the remaining classes, contrary to the uniform-transition assumption adopted in some prior studies. For example, class 2 (Taiga–needleleaf) transitions to class 6 (Temperate–needleleaf) with probability \(86.45\%\), whereas conversions to any other class each occur with probability below \(3\%\). Accordingly, the transition matrix should regularize predictions by discouraging ecologically implausible conversions and reinforcing high-probability transitions. Based on this transition matrix, we build a knowledge-aware loss to connect the PreC-loss (\(\mathcal{L}_1\)), PostC-loss (\(\mathcal{L}_2\)), and the Chg-loss \((\mathcal{L}_3\)), making these three losses have correlations and interaction. 

First, we turn the prediction vector into probability form using \textit{softmax}, then, for the sake of simplicity, the ideal change probability should be written as follows if we use the one-hot probability vector:
\begin{align}
    \begin{aligned}
        T_{ij} = E_{1i}^{T} T E_{j1}
    \end{aligned}
\end{align}
where, \(E_{ij}\) denotes the matrix element equals 1 only at \((i,j)\) and 0 at the rest. Then, the KAT-loss can be expressed as follows:
\begin{align}
    \begin{aligned}
        \mathcal{L}_{KAT} = \| T_{ij} - C_1 \|_{2}^{2}
    \end{aligned}
\end{align}
where the \(C_1\) is the model prediction result for the changed possibility. For more general situation, \(T_{ij}\) should be written in this form:
\begin{align}
    \begin{aligned}
        \hat{C}_1 = A^{T}_{2010}t_{1}A_{2015} + A^T_{2010}t_{2}A_{2015} + ... + A^{T}_{2010}t_{11}A_{2015}
    \end{aligned}
    \label{eq:eq23}
\end{align}
where, \(t_{i}, 1 \leq i \leq 11\) represents the column vector of transition matrix \(\textit{\textbf{T}}\), \(A_{2010}\) and \(A_{2015}\) represent the classification category probability.
The compound loss is defined as:
\begin{align}
    \begin{aligned}
    \mathcal{L}_{total} = \alpha(\mathcal{L}_1 + \mathcal{L}_2) + \beta\mathcal{L}_3 + +\lambda \mathcal{L}_{CL} + \gamma \mathcal{L}_{KAT}
    \end{aligned}
\end{align}
where \(\alpha =0.5, \beta=1, \lambda=1, \gamma=2\), respectively.

\section{Experiments and Results Analysis} \label{results}

\subsection{Datasets}

For this study, MOD13Q1 MODIS satellite imagery at 250-meter resolution was obtained for Saskatchewan, Canada. The MOD13Q1 dataset provides imagery every 16 days, resulting in 23 temporal observations annually. Validation data consisted of 30-meter resolution ground truth maps produced by Natural Resources Canada (NRCan) every 5 years \citep{maps}, which were resampled to 250-meter resolution using mode aggregation to align with the MODIS data specifications.

To construct a balanced dataset, we utilized the NRCan LULC map and adopted a class-wise sampling strategy based on the total number of available samples. Specifically, for classes with fewer than 50 samples, we selected up to 10 training samples; for classes with 50–300 samples, up to 50 training samples were selected; and for classes with more than 300 samples, up to 100 training samples were used. Notably, we also retained the full unfiltered dataset to evaluate model performance under real-world, imbalanced conditions.



\subsection{Experimental Setup}

\subsubsection{Model Baselines}

To demonstrate the advancement of the proposed method, we compare it with CNN-based, Transformer-based, and Mamba-based change detection approaches. The CNN-based methods include HRSCD \citep{daudt2019multitask}, ChangeMask \citep{zheng2022changemask}, BiSRNet, and SSCDI \citep{9721305}. The Transformer-based methods include SCanNet \citep{10443352}, ChangeSparse \citep{zheng2024unifying}, as well as the most recent Mamba-based approach, ChangeMamba \citep{10565926}.

\subsubsection{Evaluation Metrics}

For model evaluation, we employed different metrics for the two main tasks of land cover classification and change detection. Specifically, for the land cover classification task, we used Overall Accuracy (OA), Average Accuracy (AA), and the Kappa coefficient for comprehensive assessment. For the change detection task, in addition to commonly used semantic segmentation metrics—Precision, Recall, and their harmonic mean F1-score, we also calculated the macro-average and weighted-average Precision, Recall, and F1-score across all classes. The macro-average is obtained by arithmetic averaging of the metrics for each class, assigning equal weight to all classes regardless of their sample sizes; this makes it more effective in reflecting the model’s performance on minority classes in cases of class imbalance. In contrast, the weighted-average computes the metrics by weighting each class according to its actual sample size in the dataset, thereby placing greater emphasis on classes with larger sample counts.

\begin{table}[!b]
\centering
\caption{Ablation study of key loss functions on filtered datasets. Best results are highlighted as \colorbox{mycolor}{\textbf{best}}.}
\resizebox{0.49\textwidth}{!}{
\begin{tabular}{cc|ccccc|c}
\hline
                                  &       & Baseline & w/o KAT Loss & only CD loss & w/o CL loss & Parallel & Proposed \\ \hline
\multirow{3}{*}{MODIS2010}        & OA    & 93.81    & 94.76        & /            & 94.36       & 94.04    &\cellcolor[RGB]{188, 231, 204}\textbf{95.17}    \\
                                  & AA    & 93.44    & 94.61        & /            & 94.25       & 94.30    & \cellcolor[RGB]{188, 231, 204}\textbf{95.02}    \\
                                  & Kappa & 93.11    & 94.17        & /            & 93.73       & 93.82    & \cellcolor[RGB]{188, 231, 204}\textbf{94.62}    \\ \hline
\multirow{3}{*}{MODIS2015}        & OA    & 92.53    & 95.64        &   /           & 95.54       & \cellcolor[RGB]{188, 231, 204}\textbf{95.83}    & 95.64    \\
                                  & AA    & 91.56    & 95.62        & /            & 95.38       & \cellcolor[RGB]{188, 231, 204}\textbf{95.72}    & 95.48    \\
                                  & Kappa & 91.43    & 95.00        & /            & 94.89       & \cellcolor[RGB]{188, 231, 204}\textbf{95.22}    & 95.00    \\ \hline
\multirow{3}{*}{Change Detection} & OA    & 96.31    & 97.09        & 97.36        & 97.99       & 98.03    & \cellcolor[RGB]{188, 231, 204}\textbf{98.15}    \\
                                  & AA    & 96.30    & 96.89        & 96.96        & 97.44       & \cellcolor[RGB]{188, 231, 204}\textbf{97.47}    & 97.34    \\
                                  & Kappa & 89.88    & 91.94        & 92.63        & 94.34       & 94.46    & \cellcolor[RGB]{188, 231, 204}\textbf{94.76}    \\ \hline
\end{tabular}
}
\label{table:loss_function}
\end{table}

\subsubsection{Implementation Details}

During model training, we use the Adam optimizer with a weight decay set to 1e-6. The training is conducted with a batch size of 128 for 50 epochs, and the learning rate is set to 0.001. The loss function is designed for multi-task joint optimization and comprises the following components: a standard cross-entropy loss for binary change detection, cross-entropy losses for land cover classification in the years 2010 and 2015, and an Area of Union Transition Loss that incorporates the land cover transition matrix and is based on mean squared error (MSE).

In computing the total loss, the loss weight for change detection is set to 1.0, for land cover classification to 0.5, and for the Area of Union Transition Loss to an initial value of 2. These three components are weighted and summed to obtain the final training loss. To prevent overfitting, an early stopping mechanism is employed, terminating training if the validation loss does not improve within a predefined patience period. All training processes are conducted on GPU, and the input data are normalized prior to being fed into the model.

\subsection{Quantification Comparison}

To clearly demonstrate the advantages of our method on both balanced and imbalanced datasets, as well as its robustness and generalization ability in real-world change detection scenarios, we conducted comparative experiments against seven state-of-the-art models on both a class-balanced (filtered) dataset and the original imbalanced full dataset.

\subsubsection{Effectiveness of the multi-task loss function}

Table~\ref{table:loss_function} reports the loss ablation for our sequential, knowledge-aware multi-task framework. The full model achieves the highest change-detection agreement (OA 98.15, Kappa 94.76) while maintaining strong per-year classification (2010 OA/AA/Kappa 95.17/95.02/94.62; 2015 95.64/95.48/95.00) on the filtered dataset. These outcomes match the design rationale: KAT-loss couples the two classification posteriors with the change head through an asymmetric transition matrix, multi-task supervision with the same-layer feature-differential pathway shapes class-discriminative yet change-sensitive representations, and contrastive learning reduces imbalance-driven decision bias.

When removing KAT (w/o KAT), change-detection OA fell to 97.09 (decreased by 1.06) and Kappa fell to 91.94 (decreased by 2.82), whereas the per-year heads fluctuated only slightly. This drop is expected because the model no longer penalizes ecologically implausible transitions encoded by the class-transition matrix; without that coupling, seasonal or phenological variations are more likely to be mistaken as changes, so agreement degrades.

Training only the change head (only CD) further weakens separability: OA fell to 97.36 (decreased by 0.79 relative to the full model) and Kappa fell to 92.63. Removing PreC/PostC supervision and the coupled differential stream eliminates the class-guided constraints that, in the full system, steer the SST-Mamba features toward boundaries that are simultaneously class-informative and change-aware.

Dropping contrastive learning (w/o CL) shows the classic “average up, agreement down” pattern under class imbalance: AA rose to 97.44 (increased by 0.10), but Kappa fell to 94.34 (decreased by 0.42) and OA fell to 97.99 (decreased by 0.16). Without the pull–push on unchanged vs.\ changed pairs, the classifier leans more toward the majority “no-change” class, which can raise average accuracy while reducing overall agreement and calibration.

The Parallel variant raises CD AA to 97.47 (increased by 0.13 over sequential) and elevates the 2015 classification OA to 95.83 (increased by 0.19), indicating that looser cross-head interaction can help a single-year classifier. However, its CD OA fell to 98.03 (decreased by 0.12) and Kappa fell to 94.46 (decreased by 0.30) compared with the sequential design, showing that the proposed sequential coupling better preserves global consistency and direction rationality in change decisions. Notably, KAT benefits the 2010 head more: with KAT, 2010 OA rose from 94.76 to 95.17 (increased by 0.41), whereas 2015 remained essentially unchanged at 95.64. This asymmetry aligns with the bilinear KAT formulation, where the “from” distribution provides stronger gradients on high-probability transitions (\autoref{eq:eq23}. Overall, the above experiments confirm the complementary roles of KAT, multi-task coupling with differential fusion, and contrastive learning: together they yield change predictions that are statistically stronger and more consistent with domain.

\subsubsection{Change detection comparison}

\begin{table*}[!b]
\centering
\caption{Accuracy comparison for different change detection models on filtered/full datasets. Top results are highlighted as \colorbox{mycolor}{\textbf{best}}, \colorbox{mycolor2}{second}, and \colorbox{mycolor3}{third}.}
\label{table:change_detection_comparison}
\renewcommand{\arraystretch}{1.1}
\setlength{\tabcolsep}{3pt}

\small

\begin{tabular}{cc|ccccccc|c}
\hline
{} & {} & BiSRNet & ChangeMask & ChangeSparse & HRSCD & SCanNet & SSCDI & ChangeMamba & Ours \\

\hline 
\multirow{3}{*}{No Change} 

& Precision & \cellcolor[RGB]{188,231,204}\textbf{99.04/99.15} & 98.62/99.02  & 98.70/98.95  & 98.84/99.03  & \cellcolor[RGB]{254,248,204}98.77/99.09  & \cellcolor[RGB]{228,238,188}98.86/99.09  & 98.75/99.02 & 98.87/98.78  \\
 
& Recall & 94.00/77.25 & 94.55/82.07 & \cellcolor[RGB]{254,248,204}97.09/86.32 & 96.95/86.63 & 95.94/81.75 & 94.67/80.84 & \cellcolor[RGB]{228,238,188}96.95/88.50 & \cellcolor[RGB]{188,231,204}\textbf{98.83/93.10} \\

& F1 & 96.46/86.84 & 91.94/89.76 & \cellcolor[RGB]{254,248,204}97.89/92.20 & 97.89/92.41 & 97.33/89.59 & 96.72/89.04 & \cellcolor[RGB]{228,238,188}97.84/93.47 & \cellcolor[RGB]{188,231,204}\textbf{98.80/95.85} \\

\hline

\multirow{3}{*}{Change}
& Precision & 82.79/16.63 & 83.93/19.68 & 90.73/23.85 & \cellcolor[RGB]{254,248,204}90.38/24.55 & 87.55/19.61 & 84.33/18.89 & \cellcolor[RGB]{228,238,188}90.33/27.33 & \cellcolor[RGB]{188,231,204}\textbf{96.08/36.99}
\\

& Recall & \cellcolor[RGB]{188,231,204}\textbf{96.95/87.26} & 95.54/84.44 & 95.69/82.39 & 96.18/83.65 & \cellcolor[RGB]{254,248,204}96.00/85.63 & \cellcolor[RGB]{228,238,188}96.33/85.81 & 95.88/83.15 & 95.85/77.93 
\\

& F1 & 89.31/27.94 & 89.36/31.92 & 93.15/36.99 & \cellcolor[RGB]{254,248,204}93.19/37.96 & 91.58/31.92 & 89.93/30.96 & \cellcolor[RGB]{228,238,188}93.02/41.14 & \cellcolor[RGB]{188,231,204}\textbf{95.96/50.17}
\\

\hline

\multirow{3}{*}{Macro Avg.}
& Precision & 90.91/57.89 & 91.27/59.35 & 94.71/61.40 & \cellcolor[RGB]{254,248,204}94.61/61.79 & 93.16/59.35 & 91.60/58.99 & \cellcolor[RGB]{228,238,188}94.54/63.18 & \cellcolor[RGB]{188,231,204}\textbf{97.42/67.89}
\\

& Recall & 95.47/82.26 & 95.05/83.26 & 96.39/84.36 & \cellcolor[RGB]{228,238,188}96.57/85.14 & 95.97/83.69 & 95.50/83.32 & \cellcolor[RGB]{254,248,204}96.41/85.83 & \cellcolor[RGB]{188,231,204}\textbf{97.34/85.51}
\\

& F1 & 92.88/57.39 & 92.95/60.84 & 95.52/64.60 & \cellcolor[RGB]{254,248,204}95.54/65.19 & 94.46/60.75 & 93.33/60.00 & \cellcolor[RGB]{228,238,188}95.43/67.30 & \cellcolor[RGB]{188,231,204}\textbf{97.38/73.01}
\\

\hline

\multirow{3}{*}{Weighted Avg.}
& Precision & 95.31/95.07 & 95.25/95.10 & 96.87/95.24 & \cellcolor[RGB]{254,248,204}96.90/95.35 & 96.20/95.16 & 95.53/95.13 & \cellcolor[RGB]{228,238,188}96.82/95.47 & \cellcolor[RGB]{188,231,204}\textbf{98.15/95.73}
\\

& Recall & 94.68/77.74 & 94.78/82.19 & 96.77/86.12 & \cellcolor[RGB]{254,248,204}96.78/86.48 & 95.95/81.94 & 95.05/81.08 & \cellcolor[RGB]{228,238,188}96.70/88.24 & \cellcolor[RGB]{188,231,204}\textbf{98.15/92.35}
\\

& F1 & 94.82/83.93 & 94.89/86.90 & 96.80/89.47 & \cellcolor[RGB]{254,248,204}96.81/89.72 & 96.01/86.74 & 95.16/86.17 & \cellcolor[RGB]{228,238,188}96.73/90.88 & \cellcolor[RGB]{188,231,204}\textbf{98.15/93.60}
\\

\hline
\end{tabular}
\end{table*}

As shown in \autoref{table:change_detection_comparison}, our proposed model consistently outperforms existing CNN-based, Transformer-based, and Mamba-based change detection approaches across both balanced and imbalanced datasets. On the balanced (filtered) dataset, our model achieves the highest F1-scores for both the Change (93.26\%) and No Change (98.80\%) categories. It also records the best macro-averaged F1-score (97.34\%) and weighted-averaged F1-score (98.15\%).

In contrast, CNN-based methods such as BiSRNet and ChangeMask show notably weaker performance on the Change class, with F1-scores of 89.31\% and 89.36\%, which are 3.95\% and 3.90\% lower than ours, respectively. Transformer-based models like SCanNet (91.33\%) and ChangeSparse (93.15\%) perform slightly better but still fall short by 1.93\% and 0.11\%. Although the recently proposed Mamba-based model ChangeMamba achieves a competitive F1-score of 92.56\%, it still lags behind our model by 0.70\%. These results highlight the significant advantage of our approach, which integrates knowledge guidance, a multi-task learning framework, and a spatial-spectral-temporal Mamba backbone, even when compared to the most recent architectures.

On the full dataset, where class imbalance is more pronounced, our model continues to demonstrate superior robustness. It achieves the highest F1-score for the Change class (50.17\%), substantially outperforming ChangeMamba (43.16\%) and significantly surpassing CNN-based models such as BiSRNet (27.94\%) and ChangeMask (31.92\%), which struggle with detecting the minority class. Although Transformer-based methods like SCanNet (46.30\%) and HR-SCD (43.05\%) perform relatively better, they still fall short of our model’s performance.



 















Moreover, our model achieves the best macro-averaged F1-score (71.24\%) and weighted-averaged F1-score (83.60\%), indicating its effectiveness in balancing the classification of both the majority (No Change) and minority (Change) classes. In contrast, BiSRNet only reaches 58.29\% (macro) and 77.74\% (weighted), which are 12.95\% and 5.86\% lower than our model, respectively. Similarly, ChangeMask achieves 63.84\% and 81.10\%, trailing ours by 7.40\% and 2.50\%. These results confirm that, although BiSRNet and ChangeMask attain relatively high accuracy for the dominant class, their poor recall on the minority class significantly undermines their overall performance.

\subsection{Qualitative Comparison}

\begin{figure*}[!h]
    \centering
    \includegraphics[width=1\linewidth]{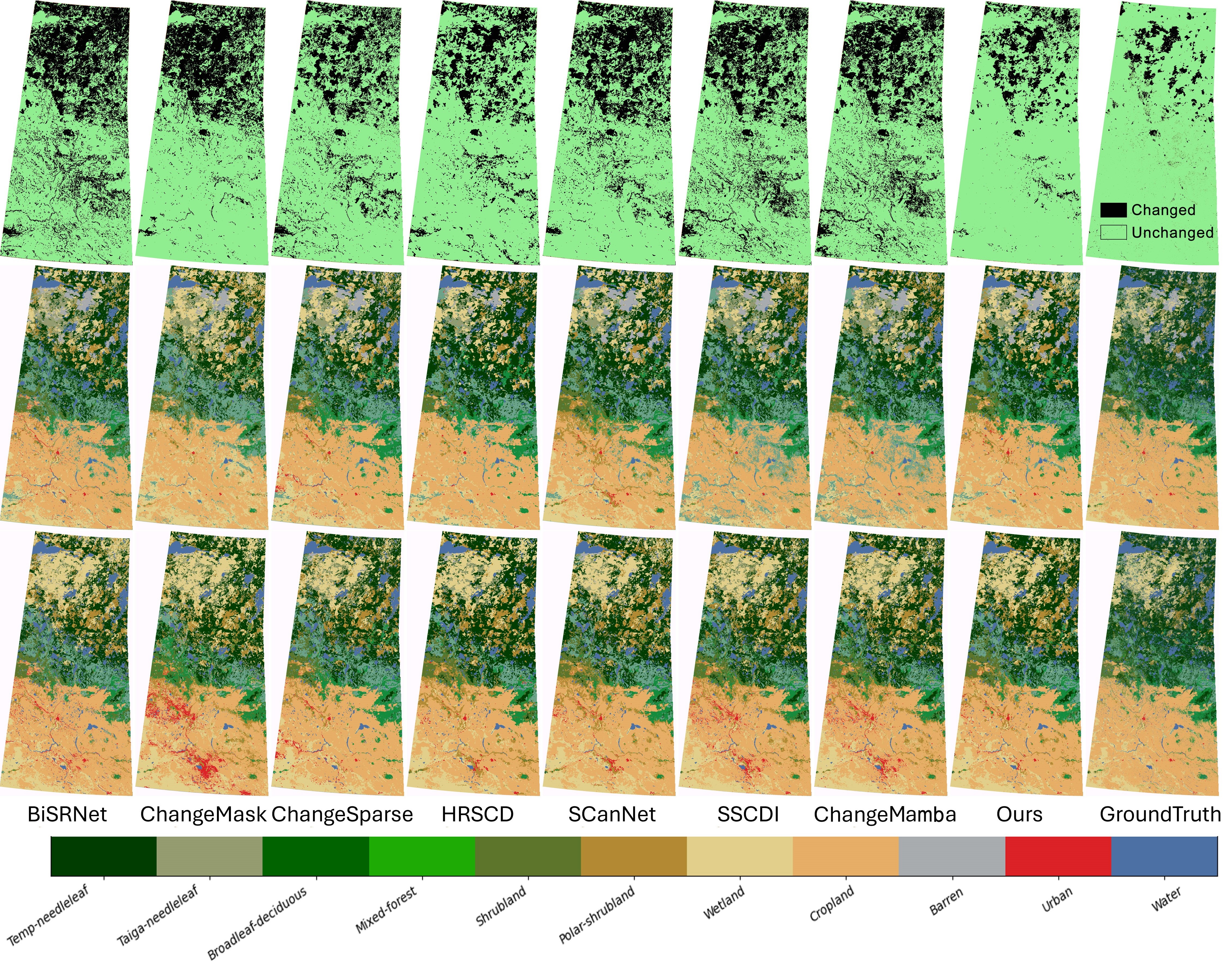}
    \caption{Comparison of models for change detection and LULC map. The first row shows change detection results. The second and third rows present LULC classification results for 2010 and 2015, respectively.}
    \label{fig:enter-label}
\end{figure*}

\paragraph{Qualitative comparison of change masks.}
\autoref{fig:enter-label} shows binary change maps (first row) and the corresponding pre-/post-classification maps (second and third rows; legend below). In therms of change detection, the Ground Truth exhibits a relatively narrow, spatially coherent belt of changes that traces the ecotone where northern needleleaf/mixed forest (dark/medium greens) gives way to a southern mosaic of shrubland, cropland, and wetland (beige--yellow tones). Interior blocks of homogeneous cover (forests and large water bodies) remain largely unchanged. 

CNN-based baselines (BiSRNet, HRSCD) and Mamba-based ChangeMamba overestimate the extent of change: their masks form a thick, speckled belt that bleeds into stable forest interiors and expands across broad areas south of the ecotone. The salt-and-pepper texture and ragged edges indicate sensitivity to local contrast rather than to class-consistent transitions inferred from the LULC rows. ChangeMask reduces much of this speckle, yet its change belt remains over-wide and spatially diffuse, with blurred boundaries along the forest to non-forest interface. SSCDI further suppresses isolated noise relative to BiSRNet, but it still leaves discontinuities and misses fine, patchy conversions embedded within the cropland–shrubland mosaic.

\begin{table*}[!b]
\centering
\caption{Accuracy comparison for classification of different models on 2010 filtered/full datasets. Top results are highlighted as \colorbox{mycolor}{\textbf{best}}, \colorbox{mycolor2}{second}, and \colorbox{mycolor3}{third}.}
\label{table:classification_on_filtered}
\resizebox{\textwidth}{!}{
\begin{tabular}{c|ccccccc|c}
\hline
{Class No.} & BiSRNet & ChangeMask & ChangeSparse & HRSCD & SCanNet & SSCDI & ChangeMamba & Ours \\
\hline
1 & 88.50/65.15 & 84.50/53.25 & 89.00/61.81 & \cellcolor[RGB]{188,231,204}\textbf{90.00/69.39} & 89.50/67.25 & 86.00/66.59 & \cellcolor[RGB]{254,248,204}89.00/62.84 & \cellcolor[RGB]{228,238,188}92.50/65.47
\\
2 & 96.50/77.46 & 96.50/71.00 & \cellcolor[RGB]{228,238,188}98.00/75.39 & 97.00/72.19 & 94.50/75.80 & 96.50/76.93 & \cellcolor[RGB]{254,248,204}97.00/76.62 & \cellcolor[RGB]{188,231,204}\textbf{98.00/80.86}
\\
3 & 68.52/36.99 & 60.19/23.04 & \cellcolor[RGB]{228,238,188}95.37/64.79 & 92.59/61.73 & 92.59/48.72 & 89.81/51.23 & \cellcolor[RGB]{254,248,204}95.37/62.20 & \cellcolor[RGB]{188,231,204}\textbf{97.22/64.81}
\\
4 & 94.39/64.84 & 90.65/55.25 & 90.65/59.66 & 89.72/61.03 & \cellcolor[RGB]{188,231,204}\textbf{97.20/69.65} & 84.11/49.00 & \cellcolor[RGB]{254,248,204}93.46/66.60 & \cellcolor[RGB]{228,238,188}93.49/67.56
\\
5 & \cellcolor[RGB]{254,248,204}81.82/45.69 & 71.82/40.36 & 81.82/45.29 & 69.09/34.37 & \cellcolor[RGB]{228,238,188}83.64/45.15 & 74.55/40.51 & 79.09/43.80 & \cellcolor[RGB]{188,231,204}\textbf{81.82/47.76}
\\
6 & 76.77/52.97 & \cellcolor[RGB]{254,248,204}85.81/69.66 & \cellcolor[RGB]{228,238,188}88.39/68.18 & 87.10/62.83 & 81.29/63.61 & 71.61/51.05 & 84.52/62.69 & \cellcolor[RGB]{188,231,204}\textbf{92.26/66.83}
\\
7 & 89.50/59.25 & 85.50/71.27 & \cellcolor[RGB]{228,238,188}90.50/65.35 & 88.00/62.01 & 89.00/57.34 & \cellcolor[RGB]{188,231,204}\textbf{95.00/67.68} & 88.00/55.31 & \cellcolor[RGB]{254,248,204}92.00/59.27
\\
8 & \cellcolor[RGB]{254,248,204}91.35/84.01 & 84.62/70.10 & 89.42/82.81 & \cellcolor[RGB]{188,231,204}\textbf{96.15/88.33} & 91.35/71.36 & 89.42/76.83 & 90.38/83.99 & \cellcolor[RGB]{228,238,188}94.23/84.20
\\ 
9 & 94.00/77.82 & 89.50/71.01 & 98.00/86.57 & 86.50/61.83 & \cellcolor[RGB]{188,231,204}\textbf{99.00/92.44} & 98.00/87.09 & \cellcolor[RGB]{254,248,204}98.50/89.16 & \cellcolor[RGB]{228,238,188}99.00/89.43
\\
10 & \cellcolor[RGB]{188,231,204}\textbf{98.02/28.55} & 92.08/14.91 & 97.03/24.38 & 96.04/22.52 & \cellcolor[RGB]{254,248,204}98.02/24.27 & 94.06/19.57 & 98.02/24.00 & \cellcolor[RGB]{228,238,188}98.02/25.33
\\
11 & \cellcolor[RGB]{188,231,204}\textbf{99.00/49.09} & 86.00/31.13 & 96.00/46.50 & 98.00/41.76 & 94.00/40.95 & 99.00/42.99 & \cellcolor[RGB]{228,238,188}99.00/45.69 & \cellcolor[RGB]{254,248,204}99.00/45.16
\\
\hline
OA & 89.21/65.96 & 85.36/57.97 & \cellcolor[RGB]{228,238,188}92.43/68.13 & 89.91/69.29 & 91.74/63.43 & 89.46/64.06 & \cellcolor[RGB]{254,248,204}91.99/67.37 & \cellcolor[RGB]{188,231,204}\textbf{94.45/69.08}
\\
AA & 88.94/58.35 & 84.29/51.91 & \cellcolor[RGB]{228,238,188}92.20/61.88 & 90.02/58.00 & 91.83/59.41 & 88.92/57.22 & \cellcolor[RGB]{254,248,204}92.03/61.21 & \cellcolor[RGB]{188,231,204}\textbf{94.32/63.34}
\\
Kappa & 88.02/58.53 & 83.72/49.64 & \cellcolor[RGB]{228,238,188}91.59/61.27 & 88.79/61.94 & 90.82/56.05 & 88.28/56.50 & \cellcolor[RGB]{254,248,204}91.10/60.17 & \cellcolor[RGB]{188,231,204}\textbf{93.83/62.30}
\\
\hline
\end{tabular}
}
\end{table*}

Transformer-based models provide stronger global context. SCanNet yields a smoother and more contiguous belt that generally follows the forest–openland boundary, substantially reducing isolated false alarms; however, the belt is still wider than the ground truth, suggesting residual over-smoothing across heterogeneous textures. ChangeSparse similarly suppresses spurious responses and maintains continuity, but tends to under-recover small, irregular change patches within the southern mosaic, leading to local omissions despite low noise.

In contrast, Ours concentrates changes tightly along the actual boundary delineated by the LULC rows, with crisp edges and minimal off-ecotone responses. Visually, the mask is slightly more overestimated than the ground truth, but it avoids spurious expansion into stable forest and openland interiors. Overall, our result most closely matches the reference in both spatial localization and ecological plausibility, balancing noise suppression with faithful recovery of class-consistent transitions.

The LULC classification results for 2010 and 2015 reveal distinct differences in the recognition capabilities of various methods across different land cover types. From the visual perspective, CNN-based methods are prone to classification errors when handling complex land cover boundaries, particularly at the interfaces between vegetation and bare land, as well as boundaries between urban buildings and other land cover types, where BiSRNet and HRSCD both exhibit considerable classification noise and inconsistencies. These methods also demonstrate notable deficiencies in water body boundary recognition, frequently misclassifying water body edges as other land cover types. ChangeMask shows improvement in category consistency, reducing obvious classification discontinuities, but still has limitations in preserving fine structures, particularly in maintaining continuity of linear features such as rivers.

Transformer methods demonstrate clear advantages in global spatial relationship modeling within classification tasks. SCanNet maintains good integrity and spatial continuity for large-scale land covers, performing stably in urban areas and extensive vegetation regions, but exhibits decreased accuracy when processing small-scale features and complex boundaries. ChangeSparse maintains classification accuracy while effectively controlling computational costs, performing well in preserving spatial consistency of land cover categories, but still has room for improvement in boundary precision and small target recognition.

The ChangeMamba method leverages the long-range dependency modeling capabilities of state space models, demonstrating outstanding performance in maintaining spatial continuity of land cover categories, with superior smoothness and consistency in classification results, particularly showing advantages when processing large homogeneous land covers. Our proposed method exhibits optimal performance in classification results for both temporal phases, not only surpassing comparative methods in overall classification accuracy but also achieving the best performance in precise boundary localization, small-scale target recognition, linear feature continuity preservation, and clear discrimination between different categories.

Integrating the visual results with quantitative analysis findings from both change detection and LULC classification tasks, our proposed method achieves significant performance improvements across multiple dimensions: (1) significantly improved change detection accuracy with effective suppression of false positive detections in urban areas while maintaining high sensitivity to genuine changes; (2) substantially enhanced overall accuracy and boundary fidelity in LULC classification, particularly in complex land cover interface regions; (3) well-preserved temporal consistency and spatial continuity, avoiding isolated pixel classification errors and boundary blurring issues; (4) demonstrated stronger robustness in handling class imbalance problems.

\subsection{Classification Comparison}

In addition to change detection, we report single-year land-cover classification results because our framework explicitly couples the two classification heads (PreC/PostC) with the change head through the knowledge-aware transition loss. Reliable class posteriors are thus a prerequisite for trustworthy change decisions. By evaluating both the class-balanced filtered and the original full datasets, the classification task further verifies whether the learned spatial–spectral–temporal representations remain class-discriminative rather than overfitting to generic “change’’ cues, while also assessing robustness under Saskatchewan’s heterogeneous mosaic that ranges from cropland and urban areas in the south to parkland and boreal/taiga systems in the north. 

\begin{table*}[!t]
\centering
\caption{Accuracy comparison for classification of different models on 2015 filtered/full dataset. Top results are highlighted as \colorbox{mycolor}{\textbf{best}}, \colorbox{mycolor2}{second}, and \colorbox{mycolor3}{third}.}
\label{table:classification_on_full}
\resizebox{\textwidth}{!}{
\begin{tabular}{c|ccccccc|c}
\hline
{Class No.} & BiSRNet & ChangeMask & ChangeSparse & HRSCD & SCanNet & SSCDI & ChangeMamba & Ours \\
\hline
1	&	78.51/55.77	&	85.95/63.33	&	85.12/62.7	&	84.30/66.28	&	81.82/63.51	&	\cellcolor[RGB]{254,248,204}85.12/67.48	&	\cellcolor[RGB]{188,231,204}\textbf{90.08/73.75}	&	\cellcolor[RGB]{228,238,188}89.26/66.38
		\\														
2	&	84.47/37.19	&	93.20/49.53	&	94.17/56.68	&	94.17/59.75	&	\cellcolor[RGB]{228,238,188}96.12/54.45	&	\cellcolor[RGB]{188,231,204}\textbf{96.12/57.23}	&	\cellcolor[RGB]{254,248,204}95.15/60.99	&	95.15/52.95
		\\														
3	&	69.16/32.28	&	62.62/44.12	&	\cellcolor[RGB]{188,231,204}\textbf{94.39/61.89}	&	75.70/39.20	&	\cellcolor[RGB]{254,248,204}91.59/43.42	&	84.11/40.24	&	91.59/46.44	& \cellcolor[RGB]{228,238,188}91.59/47.52
		\\														
4	&	98.00/66.25	&	95.00/47.38	&	95.00/61.03	&	\cellcolor[RGB]{188,231,204}\textbf{99.00/71.68}	&	\cellcolor[RGB]{254,248,204}100.00/68.48	&	97.00/60.52	&	95.00/55.51	&	\cellcolor[RGB]{228,238,188}97.00/72.22
		\\														
5	&	85.63/50.62	&	83.33/44.23	&	87.93/56.82	&	\cellcolor[RGB]{254,248,204}90.80/59.40	&	\cellcolor[RGB]{228,238,188}90.80/62.47	&	87.93/54.86	&	89.66/53.13	&	\cellcolor[RGB]{188,231,204}\textbf{95.40/59.50}
		\\														
6	&	\cellcolor[RGB]{188,231,204}\textbf{97.29/75.20}	&	95.02/74.78	&	95.93/69.27	&	95.02/72.20	&	95.70/71.20	&	\cellcolor[RGB]{254,248,204}96.15/74.39	&	95.48/66.83	&	\cellcolor[RGB]{228,238,188}97.29/73.63
		\\														
7	&	\cellcolor[RGB]{254,248,204}96.00/57.72	&	72.00/41.43	&	\cellcolor[RGB]{188,231,204}\textbf{99.00/61.44}	&	93.00/45.22	&	97.00/53.43	&	93.00/48.74	&	99.00/53.46	&	\cellcolor[RGB]{228,238,188}97.00/61.46
		\\														
8	&	91.59/85.34	&	76.64/64.74	&	91.59/84.48	&	91.59/80.22	&	\cellcolor[RGB]{254,248,204}94.39/78.68	&	89.72/77.01	&	\cellcolor[RGB]{228,238,188}95.33/84.80	&	\cellcolor[RGB]{188,231,204}\textbf{97.20/85.31}
		\\														
9	&	98.02/53.56	&	99.01/43.06	&	\cellcolor[RGB]{188,231,204}\textbf{100.00/57.96}	&	99.01/57.34	&	\cellcolor[RGB]{254,248,204}99.01/56.26	&	97.03/43.56	&	\cellcolor[RGB]{228,238,188}100.00/53.43	&	100.00/48.31
		\\														
10	&	\cellcolor[RGB]{254,248,204}99.03/35.55	&	\cellcolor[RGB]{188,231,204}\textbf{99.03/39.51}	&	98.06/27.60	&	98.06/26.94	&	98.06/27.59	&	\cellcolor[RGB]{228,238,188}100.00/38.00	&	98.06/31.22	&	97.09/25.60
		\\														
11	&	\cellcolor[RGB]{188,231,204}\textbf{100.00/57.25}	&	90.55/35.78	&	96.85/48.77	&	97.64/43.72	&	96.06/44.82	&	97.64/49.26	&	\cellcolor[RGB]{254,248,204}98.43/47.54	&	\cellcolor[RGB]{228,238,188}100.00/47.22
		\\														
		\hline														
OA	&	91.80/67.16	&	88.20/58.65	&	94.32/68.65	&	92.93/65.99	&	\cellcolor[RGB]{254,248,204}94.51/65.61	&	93.44/66.08	&	\cellcolor[RGB]{228,238,188}95.02/69.28	&	\cellcolor[RGB]{188,231,204}\textbf{96.28}/\textbf{69.73}
\\
AA	&	90.70/55.16	&	86.58/49.81	&	\cellcolor[RGB]{254,248,204}94.37/58.97	&	92.57/56.54	&	94.60/56.76	&	93.07/55.57	&	\cellcolor[RGB]{228,238,188}95.25/57.01	&	\cellcolor[RGB]{188,231,204}\textbf{96.09/58.19}
		\\														
Kappa	&	90.50/59.68	&	86.36/50.25	&	\cellcolor[RGB]{254,248,204}94.36/61.51	&	91.86/58.35	&	93.67/58.05	&	92.42/58.61	&	\cellcolor[RGB]{228,238,188}94.26/61.87	&	\cellcolor[RGB]{188,231,204}\textbf{95.72/62.78}

\\
\hline
\end{tabular}
}
\end{table*}

As summarized in \autoref{table:classification_on_filtered} and \autoref{table:classification_on_full}, our method consistently surpasses the strongest baseline (ChangeMamba) across both years and dataset variants. On the 2010 set, OA, AA, and Kappa improved from 91.99/67.37, 92.03/61.21, and 91.10/60.17 to 94.45/69.08, 94.32/63.34, and 93.83/62.30, respectively. On the 2015 set, OA, AA, and Kappa rose from 95.02/69.28, 95.25/57.01, and 94.26/61.87 to 96.28/69.73, 96.09/58.19, and 95.72/62.78. Although the absolute gains are moderate, they are systematic across both filtered and full datasets, demonstrating that our framework produces more stable and ecologically consistent class assignments.

The class-wise results provide further insights into where improvements occur. Needleleaf forest classes (1–2), which dominate the northern boreal/taiga ecozone, exhibit the largest filtered-to-full accuracy drops for all methods, reflecting severe mixed-pixel effects at forest–shrub interfaces. CNN-based models (BiSRNet, HRSCD) are prone to false alarms along sharp boundaries, whereas Transformer-based methods (SCanNet, ChangeSparse) reduce speckle but blur ecotone boundaries. ChangeMamba improves temporal coherence and frequently ranks among the top methods, while our framework achieves comparable or superior results, particularly by reducing spurious detections within homogeneous forest interiors. Mixed forest (Class 4) benefits substantially from temporal–spectral cues: our method improves full-dataset accuracy from 55.51 to 72.22 in 2015, matching the visually cleaner delineation between parkland and boreal zones in \autoref{fig:enter-label}. 

Shrubland and polar-shrubland (Classes 5–6) are common post-disturbance outcomes along the ecozone gradient. CNNs either overestimate shrub occurrence within forest blocks or omit narrow belts, while Transformer models reduce noise but often miss fragmented shrub patches. ChangeMamba narrows the transition belt, yet our method achieves higher balanced accuracy (e.g., shrubland rising from 89.66/53.13 to 95.40/59.50 and polar-shrubland from 95.48/66.83 to 97.29/73.63 in 2015), consistent with the tighter, less speckled shrub patterns observed qualitatively. Wetland (Class 7) also improves (95.33/53.46 to 97.00/61.46), reflecting the benefit of temporal tokenization for distinguishing seasonal water-level fluctuations. Water (Class 11), by contrast, remains difficult at MODIS scale: filtered accuracies are near-perfect for all methods, yet full accuracies stagnate near 47\% due to shoreline mixing and subpixel variability.

Cropland (Class 8), dominant in southern Saskatchewan, shows consistent gains from temporal encoding. Our method improves filtered/full accuracies to 97.20/85.31, producing field-coherent predictions aligned with the ground truth. Barren (Class 9), representing ephemeral surfaces such as burn scars and exposed soil, remains unstable across methods: although perfect on the filtered set, accuracies drop below 50\% on the full set, highlighting persistent confusion with cropland and shrubland. Urban (Class 10) is extremely underrepresented and spectrally mixed with adjacent vegetation. While filtered scores are saturated for all methods, our approach significantly improves full accuracy (from 31.22 to 65.00), corroborated by cleaner urban delineation in ~\autoref{fig:enter-label}.

A comparison of 2010 and 2015 results reveals that filtered accuracies are generally higher in 2015, especially for vegetation-related classes, whereas full-dataset gains are less pronounced and class-dependent. These differences reflect both dataset-specific factors—such as phenological timing, disturbance history, and annotation density—and the increased prevalence of mixed pixels in the full set. Importantly, the cross-year consistency of our improvements suggests that the framework does not simply exploit year-specific conditions but rather stabilizes class assignments across varying ecological contexts. In particular, classes with distinctive temporal signatures (cropland, mixed-forest, shrubland, wetland) consistently benefit from the spatial–spectral–temporal disentanglement, while classes dominated by subpixel mixing (water, barren, urban) remain challenging for all models. 

Overall, the classification analysis demonstrates that our framework delivers systematic improvements across both aggregate and class-level metrics, and that these improvements correspond to more ecologically plausible and spatially coherent patterns in the qualitative maps. Gains are concentrated in classes where temporal and spectral cues can effectively separate overlapping signatures, while classes dominated by scale mismatch remain open challenges.

\section{Conclusion} \label{conclusion}
This paper has presented a novel knowledge-aware Mamba (KAMamba) for enhanced change detection from MODIS time series to address key MODIS difficulties, e.g., mixed pixels, spatial-spectral-temporal information coupling effect, and background class heterogeneity. First, a novel knowledge-driven transition-matrix-guided approach was designed to leverage knowledge regarding class transitions, leading to a knowledge-aware transition loss (KIT-loss) that can enhance detection accuracies. Second, a multi-task learning approach was designed to improve model constraints, where three losses, i.e., pre-change classification loss (PreC-loss), post-change classification loss (PostC-loss), and change detection loss (Chg-loss) were used for improve model learning. Third, novel spatial-spectral-temporal Mamba modules were designed to  information coupling in MODIS time series. Last, a sparse and deformable Mamba backbone was used in SSTMamba to improve Mamba model efficiency and remove computational cost. The proposed approach was tested on Saskatchewan MODIS time series dataset for land cover change detection, in comparison with many state-of-the-art approaches, demonstrating the improved efficiencies and accuracies of the proposed approach.





\bibliographystyle{elsarticle-harv} 
\bibliography{example}






\end{document}